\documentclass{article}

\usepackage[utf8]{inputenc} 
\usepackage[T1]{fontenc}    
\usepackage{hyperref}       
\usepackage{url}            
\usepackage{booktabs}       
\usepackage{amsfonts}       
\usepackage{amsmath}
\usepackage{nicefrac}       
\usepackage{microtype}      
\usepackage{xcolor}         
\usepackage{multirow}
\usepackage{graphics}
\usepackage{algorithm}
\usepackage{algpseudocode}
\usepackage{graphicx}
\begin{document}

\title{FinNet: Finite Difference Neural Network for Solving Differential Equations}

\author{
Son N. T. Tu*\thanks{* denotes equal contribution}\footnote{University of Wisconsin Madison, USA}
\qquad
Thu Nguyen*\footnote{SimulaMet, Norway}
}
\maketitle

\begin{abstract} 
Deep learning approaches for partial differential equations (PDEs) have received much attention in recent years due to their mesh-freeness and computational efficiency. However, most of the works so far have concentrated on time-dependent nonlinear differential equations. In this work, we analyze potential issues with the well-known Physic Informed Neural Network for differential equations with little constraints on the boundary (i.e., the constraints are only on a few points). This analysis motivates us to introduce a novel technique called FinNet, for solving differential equations by incorporating finite difference into deep learning. Even though we use a mesh during training, the prediction phase is mesh-free. We illustrate the effectiveness of our method through experiments on solving various equations, which shows that FinNet can solve PDEs with low error rates and may work even when PINNs cannot. 
\end{abstract}

\section{Introduction}
Differential equations play a crucial role in many aspects of the modern world, from technology to supply chain, economics, operational research, and finance \cite{Han8505}. Solving these equations numerically has been an extensive area of research since the first conception of the modern computer. Yet, there are some potential drawbacks of classical methods, such as finite difference and finite element. Firstly, the \emph{curse of dimensionality}, that is, the computational cost, increases exponentially with the dimension of the equation \cite{hutzenthaler2020overcoming}. Secondly, classical methods usually need a mesh \cite{zang2020weak,ren2020higher}.
With the advancement of deep learning, there have been many works on using neural networks to solve differential equations that potentially can shed light on resolving the above difficulties~\cite{ruthotto2020deep,kumar2011multilayer}.


One of the foundational works in deep learning for solving partial differential equations is PINNs \cite{raissi_physics_2017}. Here, a neural network is
trained to solve supervised learning tasks with respect to given laws of physics described by the nonlinear partial differential equations. Various variants or extension of this method exist. For example, XPINNs \cite{jagtap2020extended} is a generalized space-time domain decomposition framework for PINNs to solve nonlinear PDEs in arbitrary complex-geometry domains. Another example is PhyGeoNet \cite{gao2021phygeonet}, a CNN-based variant of PINNs for solving PDEs in an irregular domain. 

In another work\cite{Han8505}, the authors try to address the curse of dimensionality in high-dimensional semi-linear parabolic PDEs by reformulating the PDEs using backward stochastic differential equations and approximating the gradient of the unknown solution by deep reinforcement learning with the gradient acting as the policy function. 
Further notable work on high-dimensional PDEs is Deep Galerkin Method \cite{sirignano2018dgm}, in which the solution is approximated by a neural network trained to satisfy the differential operator, initial condition, and boundary conditions using batches of randomly sampled time and space points. In addition, the authors in \cite{grohs2020deep} consider using deep neural network for high-dimensional elliptic PDEs with boundary conditions. 

Furthermore, SPINN \cite{ramabathiran2021spinn} is a recently developed method that uses an interpretable sparse neural network architecture for solving PDEs and the authors in \cite{dung2021deep} propose a deep ReLU neural network approximation of parametric and
stochastic elliptic PDEs with lognormal inputs.

However, most of the works in the field of deep learning for differential equations are for time-dependent partial differential equations \cite{raissi_physics_2017,jagtap2020extended, sirignano2018dgm, Han8505}. Therefore, it would be interesting to explore how deep learning techniques can be used in other scenarios. In this work, we illustrate via examples that applying PINNs to certain PDEs may not give desirable results. We investigate potential reasons for such problems and propose a novel method, namely Finite Difference Network (FinNet), that uses neural networks and finite difference to solve such equations. 

The main contributions of this work are the following: (1) We show examples we PINNs fails to work for PDEs with very few constraints on the boundary and analyze the potential reason; (2) We propose FinNet, a method based on finite difference and neural network to solve PDE with little constraints on the boundary;  (3) We illustrate via various examples that FinNet can solve PDEs efficiently, even when PINNs cannot; (4) We discuss open problems for future research. 

The rest of the paper is organized as follows: 
First, section \ref{sec: prelim} gives some preliminaries on PINNs for solving time-dependent nonlinear partial differential equations (PDEs). In section \ref{sec:motiv}, we explore the potential issues with applying PINNs for some differential equations that are not time-dependent nonlinear, analyze the examples, and give motivation to our FinNet approach. Next, section \ref{sec:method} details our FinNet method, and section \ref{sec: example} gives various examples on applying FinNet to solve differential equations. Lastly, the paper ends with a conclusion of this work and open questions in section \ref{sec:concl}. 


\section{Preliminaries: Physics Informed Neural Networks} \label{sec: prelim}
PINNs \cite{raissi_physics_2017} considers parameterized and nonlinear partial differential equations of the form 
    $u_t + \mathcal{N} [u; \lambda ]=0,$ 
where $u(t,x)$ is the latent solution, and $ \mathcal{N} [.; \lambda ]$ is a nonlinear operator parameterized by $\lambda$. 
It defines
    $f := u_t + \mathcal{N}[u],$
and approximates $u(t,x)$ by a neural network.
Then, the parameters of the neural network and $f(t,x)$ can be learned by minimizing the mean squared error (MSE) loss
\begin{equation}\label{msePINNS}
    L = \frac{1}{N_u}\sum_{i=1}^{N_u} |u(t^i_u, x^i_u)-u^i|^2 + \frac{1}{N_f} \sum_{i=1}^{N_f} |f(t_f^i,x_f^i)|^2,
\end{equation}
where $\{t_u^i, x_u^i, u^i\}_{i=1}^{N_u}$ is the initial and boundary training data on $u(t,x)$ and $\{t_f^i, x_f^i\}_{i=1}^{N_f}$ is the collocations points for $f(t,x)$.

For example, consider solving the Burger equation with Dirichlet boundary conditions
\begin{equation}
\left\{\begin{matrix}
u_t+uu_x-(0.01/\pi)u_{xx}&=0,\;\; x\in [-1,1],\;\; t\in [0,1],\\
    u(0,x) = -sin(\pi x),\;\;\;\;\;\;&\\
     u(t,-1) = u(t,1) = 0.\;\;\;&
\end{matrix}\right.
\end{equation}
Then, PINNs defines
\begin{equation}
    f = u_t + uu_x - (0.01/\pi) u_{xx},
\end{equation}
and approximate $u(t,x)$ by a neural network. Next, the parameters of the neural network $u(t,x)$ can be learned by minimizing the MSE:
\begin{equation}\label{eq4}
    L = \frac{1}{N_u}\sum_{i=1}^{N_u} |u(t^i_u, x^i_u)-u^i|^2 + \frac{1}{N_f} \sum_{i=1}^{N_f} |f(t_f^i,x_f^i)|^2, 
\end{equation}
where $\{t_u^i, x_u^i, u^i\}_{i=1}^{N_u}$ is the initial and boundary training data on $u(t,x)$ and $\{t_f^i, x_f^i\}_{i=1}^{N_f}$ is the collocations points for $f(t,x)$.

\section{Motivation}\label{sec:motiv}
In this section, we first illustrate via examples that, in some cases, applying PINNs to solve differential equations may not lead to convergence towards the desired solution.
We attempt to explain potential reasons why such an issue can arise and by this, provides motivation for our approach.

\subsection{Example 1: }\label{sec:ex1}
Consider the following equation 
\begin{align}\label{eqex1}
\begin{cases}
u'(x) + u(x) = x, \text{ for}\; 0<x<1,\\
u(0) = 1.
\end{cases}
\end{align}
The exact solution is
\begin{equation}
    u^*(x) = x-1+2e^{-x}.
\end{equation}
To solve this equation by PINNs, we approximate $u$ by a neural network with $4$ layers, each layer has $32$ neurons, and \textit{tanh} as activation function. We train the network with $5,000$ epochs and the following loss function 
\begin{equation}
    L = \frac{1}{99}\sum_{i=1}^{99} \left( \frac{d\hat{u}}{dx_i} + \hat{u} -x_i\right)^2 + |\hat{u}(0)-1|^2
\end{equation}
Here, $x_1=0.01, x_2 = 0.02,..., x_{99}=0.99$ is the training data.

\begin{figure}[htbp]
    \centering
    \includegraphics[scale=0.5]{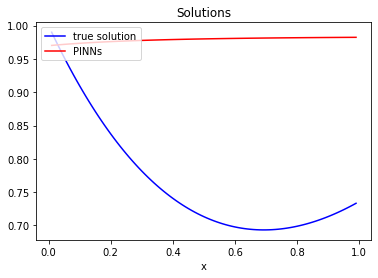}
    \includegraphics[scale=0.5]{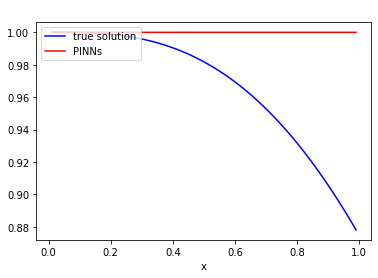}
    \caption{Left: Approximation by PINNs compared to the true solution for equation \ref{eqex1}. Right: The true solution versus the neural network solution for equation \ref{ex2pinns}}
    \label{figex1pinns}
\end{figure}

After $5,000$ epochs, the loss becomes as low as $5.15\times 10^{-5}$. Yet, figure \ref{figex1pinns} (left figure) shows that the approximation from the neural network is not close to the true solution. Examining the gradients shows that $u'(x) \approx 0.0125$ at all interior points (the mean of $u'(x_i),i=1,2,...,n$ is $0.0125$ and the variance is $0.0001$). 


\subsection{Example 2: second order static equation}\label{sec:ex2}
 We attempted to solve the following initial boundary equation 
\begin{equation}\label{ex2pinns}
\begin{cases}
u''(x) +u(x) = e^{-x}, \qquad\text{for}\; 0<x<1,\\
u(0) = 1,  \qquad u\left(1\right) = \frac{1}{2}\cos(1) + \frac{1}{2}\sin(1) + \frac{e^{-1}}{2}.
\end{cases}
\end{equation}

The exact solution (viscosity solution) is
\begin{equation}
    u^*(x) = \frac{\cos x+\sin x + e^{-x}}{2}.
\end{equation}

In an attempt to solve this equation by PINNs, we approximate $u$ using a neural network with $4$ layers, where each layer has $32$ neurons and a \textit{tanh} activation function. We train the network with $5,000$ epochs and the following loss function:
\begin{align*}
     L &= \frac{1}{99}\sum_{i=1}^n\left(\frac{d^2\hat{u}}{dx_i^2} + \hat{u}(x_i) - e^{-x_i}\right)^2 
     + \frac{1}{2}\left(|\hat{u}(0)-1|^2 + \left|\hat{u}(1)-\frac{1}{2}\cos(1) + \frac{1}{2}\sin(1) + \frac{e^{-1}}{2}\right|^2\right).
\end{align*}
Here, $x_1=0.01, x_2 = 0.02, ..., x_{99}=0.99$ is the training data. The approximated solution produced by PINNs is provided in figure \ref{figex1pinns} (right figure).


After $50$ epochs, the loss reduces to  $2.88$ and then stays approximately the same throughout epoch $51$ to epoch $5,000$. From figure \ref{figex1pinns}, we can see that the approximation from the neural network is almost constant rather than being close to the true solution. Examining the gradients shows that $u''(x) \approx 0$ at all interior points (the mean of $u'(x_i),i=1,2,...,n$ is $-7.86\times 10^{-5} $ and variance is $9.53\times 10^{-9}$). Hence, we can say that the neural network gets stuck at a local minima in this case.

\subsection{Analysis and Motivation for FinNet} \label{ana}

By the \emph{Universal approximation theorem} for neural network (\cite{LESHNO1993861, HORNIK1991251}), PINNs' approximation is always possible given enough parameters. However, from the examples above, we see that applying PINNs to certain kinds of differential equations may not give a desirable result, and the network may get stuck at a local minimum. However, note that training in this manner does not involve any label, and PINNs seems to work well for nonlinear time-dependent PDEs as studied in \cite{raissi_physics_2017}. 
Further, without boundary constraints, a PDE fails to have a unique solution. In addition,  when training a neural network to solve a differential equation, we need to inform the network about the constraint on the boundary. Next, recall that in equation \ref{eq4}, the constraints on the boundary is informed to the network via the term $\frac{1}{N_f} \sum_{i=1}^{N_f} |f(t_f^i,x_f^i)|^2$, which is based on $N_f$ points. For a time-dependent equation, $N_f$ can be reasonably large and feed into the network enough information for convergence to a desirable result. However, for the PDEs in equation \ref{eqex1} and equation \ref{ex2pinns}, the boundary consists of only two points.  

This motivates us to provide more instructions for the neural network learning process by incorporating the finite difference mechanism into the network, which informs the network that the data points should satisfy the conditions stated by finite difference. In addition, $u(x,y)$ is known at the boundary. For example, in Equation \ref{ex2}, the boundary is known to be 
\begin{equation}
    u(0) = 1,  \qquad u\left(1\right) = \frac{1}{2}\cos(1) + \frac{1}{2}\sin(1) + \frac{e^{-1}}{2}.
\end{equation}
Therefore, instead of minimizing the MSE as in equation \ref{msePINNS}, we will use this information along with finite difference to estimate the derivative terms. This helps estimate derivatives at the boundary more accurately and provides the learning process with better instructions on what the network should satisfy. The method will be presented in the next section.

\section{Finite Difference Network (FinNet)}\label{sec:method}
This section details our finite difference network (FinNet) approach. Assume that we have a function $f: \mathbb{R} \rightarrow \mathbb{R}$, and a (uniform) mesh $..., x_{i-2},x_{i-1},x_i, x_{i+1},x_{i+2},... $ with $h = x_{i+1} - x_i$. Then, recall that by using finite difference, the first order derivative $f'(x_i)$ can be computed approximately by one of the following three formulas
\begin{equation} 
    f'(x_i)\approx \frac{f(x_{i+1}) - f(x_i)}{h},   \;
    f'(x_i)\approx \frac{f(x_{i+1}) - f(x_i)}{h}, \;
    f'(x_i)\approx \frac{f(x_{i+1}) - f(x_{i-1})}{2h},
\end{equation}
and the second order derivative can be approximated by
\begin{equation}
    \frac{f(x_{i+1}) - 2f(x_i) + f(x_{i-1})}{h^2}.
\end{equation}
and for the general case where $f: \mathbb{R}^n\rightarrow \mathbb{R}$ then the derivative terms are estimated by using the above univariate finite difference scheme to the partial derivatives of $f$.

Next, we define some definitions and notations in table \ref{tab:notations}.
\begin{table}[htbp]
\caption{Table of Notations}
\begin{center}
\begin{tabular}{|c|c|}
\hline
\textbf{Notations}&\textbf{Descriptions}\\
\hline
$\Omega$& an open subset of $\mathbb{R}^n$ \\
$\partial \Omega$ & the boundary of $\Omega$\\
$G$ & a set of meshgrid points \\
$B$ & a set of boundary points, $B\subset G$ \\
$u^*$ & the true solution \\
$v$ & a neural network that approximate $u^*$\\
$L$ & loss function \\
$N$ & mesh grid size\\
$MSE(a,b)$ & mean squared error between vector $a$ and vector $b$\\
\hline
\end{tabular}
\label{tab:notations}
\end{center}
\end{table}

For a continuous operator $F$, to solve the following problem
\begin{equation}\label{eqf}
    \begin{cases}
        F(x,u,Du,D^2u) = 0 \qquad\text{in}\;\Omega,\\
        u = g\qquad\text{on}\;\partial\Omega,
    \end{cases}
\end{equation}
We discretize $[a,b] = \{x_1,x_2,\ldots, x_N\}$ and for simplicity we use uniform mesh size $\Delta = (N+1)^{-1}(b-a)$ as the distance between two consecutive points.


 The FinNet strategy for solving differential equations is as given in Algorithm \ref{alg:finnet}. Given a neural network model $v$, we train the network as following: For each epoch, we first compute $\hat{u} \gets v(G), \;\hat{u}_B \gets v(B)$. Note that $B\subset G$ so the computation of $\hat{u}_B \gets v(B)$ is already done in the $\hat{u} \gets v(G)$ operation. Though, we write it down to the clarity of the $\hat{u}_B$ notation. Then, we initialize the loss $L$ with the MSE loss at the boundary: $L \gets MSE(\hat{u}_B, g(B))$. This is to ensure that the constraint $u=g$ on $\partial \Omega$ is satisfied. Next, we update the boundary values of $\hat{u}$ with the already known exact values based on $u=g$ on $\partial \Omega$ as in equation \ref{eqf}. This is done by assigning $\hat{u}_B \gets g(B)$. Based on this newly updated $\hat{u}$, we estimate the derivatives in $F$ by finite difference. This later allows us to estimate $F$ based on the approximated terms. Then, we update the loss: $L \gets L +  MSE (F(x,u,\hat{D}u,\hat{D} ^2u), 0)$. This is to ensure that the condition $F(x,u,Du,D^2u) = 0\; \text{in}\;\Omega$ is satisfied. After getting the loss, we update the weights of the neural network $v$.

Note that the step "update the boundary values of $\hat{u}$ with the already known exact values based on $u=g$ on $\partial \Omega$ as in equation \ref{eqf}. This is done by assigning $\hat{u}_B \gets g(B)$." is crucial. Estimating the derivatives terms by Finite Difference using this is more accurate than using the predicted values of the network on the boundary.
\begin{algorithm}
\caption{FinNet }\label{alg:finnet}
\textbf{Input:} 
\begin{itemize}
    \item 
a PDE to solve:
\begin{equation}\label{eqfalg1}
    \begin{cases}
        F(x,u,Du,D^2u) = 0 \qquad\text{in}\;\Omega,\\
        u = g\qquad\text{on}\;\partial\Omega.
    \end{cases}
\end{equation}
\item $v$: a neural network to approximate $u$,

\item a set of meshgrid points $G$, a set of boundary points $B\subset G$,
\item $\hat{D} ^iu:$ an approximation of $D^iu$ by finite difference, $i=1,2$.
\end{itemize}
\textbf{Training $v$:}

\begin{algorithmic}

\For{\textbf{e} in epochs}
    \State $\hat{u} \gets v(G), \;\hat{u}_B \gets v(B)$
    \State $L \gets MSE(\hat{u}_B, g(B))$
    \State {Update the boundary values of} $\hat{u}$ by setting: $\hat{u}_B \gets g(B)$
    \State Estimate the derivatives in $F$ by finite difference based on $\hat{u}$
    \State $L \gets L + MSE (F(x,u,\hat{D}u,\hat{D} ^2u), 0)$
    \State {Update} the weights of $v$
\EndFor
\State \Return $v$
\end{algorithmic}
\end{algorithm}
Another noteworthy point is that since we use finite difference during the training phase, a mesh is needed at this stage. However, similar to PINNs, the prediction phase is mesh-free.

\section{Examples} \label{sec: example}
In this section, we provide various examples on how FinNet can be use to solve differential equations. The source code for the examples will be made available upon the acceptance of the paper.

\subsection{Example 1: Linear first-order equation}
Again, consider the equation \ref{eqex1} in section \ref{sec:motiv}
    \begin{equation}\label{ex1eq}
    \left\{\begin{matrix}
    u'(x) + u(x) &= x, \qquad\; 0<x<1,\\
    u(0) &= 1.
\end{matrix}\right. 
    \end{equation}
    
The true solution is
     $u^*(x) = x-1+2e^{-x}.$
For this equation, we let
\begin{equation}
    F = u'(x) + u(x) - x.
\end{equation}

We used a neural network of two hidden layers with 16 neurons/layer and hyperbolic tangent activation functions to approximate the true solution. To learn the parameters, we use the Adam optimizer with learning rate $0.01$. In this case,
    $G = \{0,0.01,0.02,..., 0.99, 1\},
    B = \{0, 1\}.$

Following the FinNet strategy, we train the network $v$ with the optimization in each epoch as follows, we first compute $\hat{u} \gets v(G)$, which also gives $\hat{u}(0) = v(0)$. Then, we initialize 
    $L \gets |\hat{u}(0) -1|^2$
to enforce the boundary constraint $u(0)=1$ on the neural network. Next, we update the boundary values of $\hat{u}$ with the already known exact values, i.e., update
    $\hat{u}(0) \gets 1.$ 
 Based on this newly updated $\hat{u}$, we estimate the derivatives $u_{xx}, u_{yy}$ by finite difference. Then, we update the loss:
\begin{equation}
    L \gets L + \frac{1}{99}\sum_{i=1}^{99} \left(\left\vert \hat{u}'(x_i) + u(x_i) -x_i\right\vert^2\right),
\end{equation}
where $x_1 = 0.01, x_2 = 0.02,..., x_{99}=0.99$. After getting the loss, we update the weights of the neural network $v$.

\begin{figure}[htbp]
\centering
\includegraphics[scale=0.45]{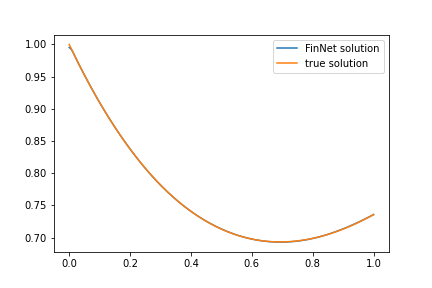}
\includegraphics[scale=0.45]{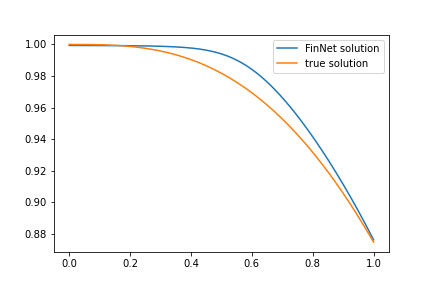}
    \caption{Left: True solution versus the neural network's approximated solution for equation \ref{ex1eq}. Right: True solution versus the neural network's approximated solution for equation \ref{ex2}.}
    \label{figex1}
\end{figure}

After $5,000$ epochs, the loss goes down to $3.34\times 10 ^{-5}$, and the mean square error between the true solution and the predicted values is  $1.15\times 10^{-7}$. The plot of the true solution versus the neural network's approximated solution is as shown in figure \ref{figex1} (left figure).

\subsection{Example 2: Second-order linear equation}
Consider the following initial boundary equation, which we have tried to solve by PINNs in section \ref{sec:motiv},
\begin{equation}\label{ex2}
\begin{cases}
u''(x) +u(x) = e^{-x}, \qquad\text{for}\; 0<x<1,\\
u(0) = 1,  \qquad u\left(1\right) = \frac{1}{2}\cos(1) + \frac{1}{2}\sin(1) + \frac{e^{-1}}{2}.
\end{cases}
\end{equation}

The exact solution (viscosity solution) is
\begin{equation}
    u^*(x) = \frac{\cos x+\sin x + e^{-x}}{2}.
\end{equation}

In this case, 
\begin{equation}
    F = u''(x) +u(x) - e^{-x}.
\end{equation}

We used a neural network consisting of $2$ hidden layers with $16$ neurons per layer and hyperbolic tangent activation functions to approximate the true solution. To learn the parameters, we use the Adam optimizer \cite{kingma2014adam} with learning rate $0.01$. In this case,
\begin{align*}
    G &= \{0,0.01,0.02,..., 0.99, 1\},\\
    B &= \{0, 1\}.
\end{align*}

Following the FinNet strategy, we train the network $v$ with the optimization in each epoch as follows, we first compute $\hat{u} \gets v(G)$, which also gives $\hat{u}(0) = v(0)$ and $\hat{u}(1) = v(1)$. Then, we initialize 
\begin{equation}
    L \gets \frac{1}{2}\left(|\hat{u}(0) -1|^2 
     + \left|\hat{u}(1) -\left(\frac{1}{2}\cos(1) + \frac{1}{2}\sin(1) + \frac{e^{-1}}{2}\right)\right|^2\right)
\end{equation}
to enforce the boundary constraints on the neural network. Next, we update the boundary values of $\hat{u}$ with the already known exact values, i.e., update
\begin{equation}
    \hat{u}(0) \gets 1,  \qquad \hat{u}\left(1\right) \gets \frac{1}{2}\cos(1) + \frac{1}{2}\sin(1) + \frac{e^{-1}}{2}.
\end{equation}
 Based on this newly updated $\hat{u}$, we estimate the derivatives $u''(x_i)$ by finite difference. Then, we update the loss:
\begin{equation}
    L \gets L + \frac{1}{99}\sum_{i=1}^{99} (u''(x_i) +u(x_i) - e^{-x_i}),
\end{equation}
where $x_1 = 0.01, x_2 = 0.02,..., x_{99}=0.99$. After getting the loss, we update the weights of the neural network $v$.

After $5,000$ epochs, the loss goes down to 0.733, and the MSE is between the true solution and the predicted values is  $4.99\times 10^{-5}$. The plot of the true solution versus the neural network's approximated solution is as shown in figure \ref{figex1} (right picture).

\subsection{Example 4: Laplace equation in two dimension}

Let $\Omega =  (-1,1)\times (-1,1)$, the problem is
\begin{equation}\label{laplaceeq}
\begin{cases}
u_{xx}+u_{yy} = 0, & \qquad\text{for}\; (x,y)\in (-1,1)\times (-1,1),\\
u(x,y) = xy , &\qquad\text{for}\; (x,y)\in \partial \Omega.
\end{cases}
\end{equation}
The exact solution is 
    $u^*(x,y) = xy.$

We used a neural network $v$ of two hidden layers with 8 neurons per layer and hyperbolic tangent activation functions to approximate the true solution. To learn the parameters, we use the Adam optimizer with learning rate 0.01. The mesh size used is $N=32$. Then $G$ is formed by the grid $[-1,\frac{-29}{31},...,\frac{29}{31},1]\times [-1,\frac{-29}{31},...,\frac{29}{31},1]$ and $B$ consists of boundary grid points $(-1, -1), (-1, \frac{-29}{31}),..., (-1,\frac{29}{31}),(-1,1)$ and $(1, -1), (1, \frac{-29}{31}),..., (1,\frac{29}{31}),(1,1)$. 

Following the FinNet strategy, we train the network $v$ as follows: For each epoch, we first compute $\hat{u} \gets v(G)$, which also gives $\hat{u}_B \gets v(B)$. Then, we initialize $L \gets MSE(\hat{u}_B, g(B))$ to enforce the boundary constraint $u(x,y)=xy$ on $\partial \Omega$ on the neural network. Next, we update the boundary values of $\hat{u}$ with the already known exact values based on $u(x,y)=xy$ for $(x,y)\in B$. Based on this newly updated $\hat{u}$, we estimate the derivatives $u_{xx}, u_{yy}$ by finite difference. Then, we update the loss:
\begin{equation}
    L \gets L + \frac{1}{30^2}\sum_{i=1}^{30}\sum_{j=1}^{30}(u_{x_ix_i}+u_{y_jy_j}),
\end{equation}
where $x_1 = \frac{-29}{31}, ..., x_{30} =  \frac{29}{31}$ and $y_1 = \frac{-29}{31}, ..., y_{30} =  \frac{29}{31}$. After getting the loss, we update the weights of the neural network $v$.

\begin{figure}[htbp]
\centering
\includegraphics[scale=0.06]{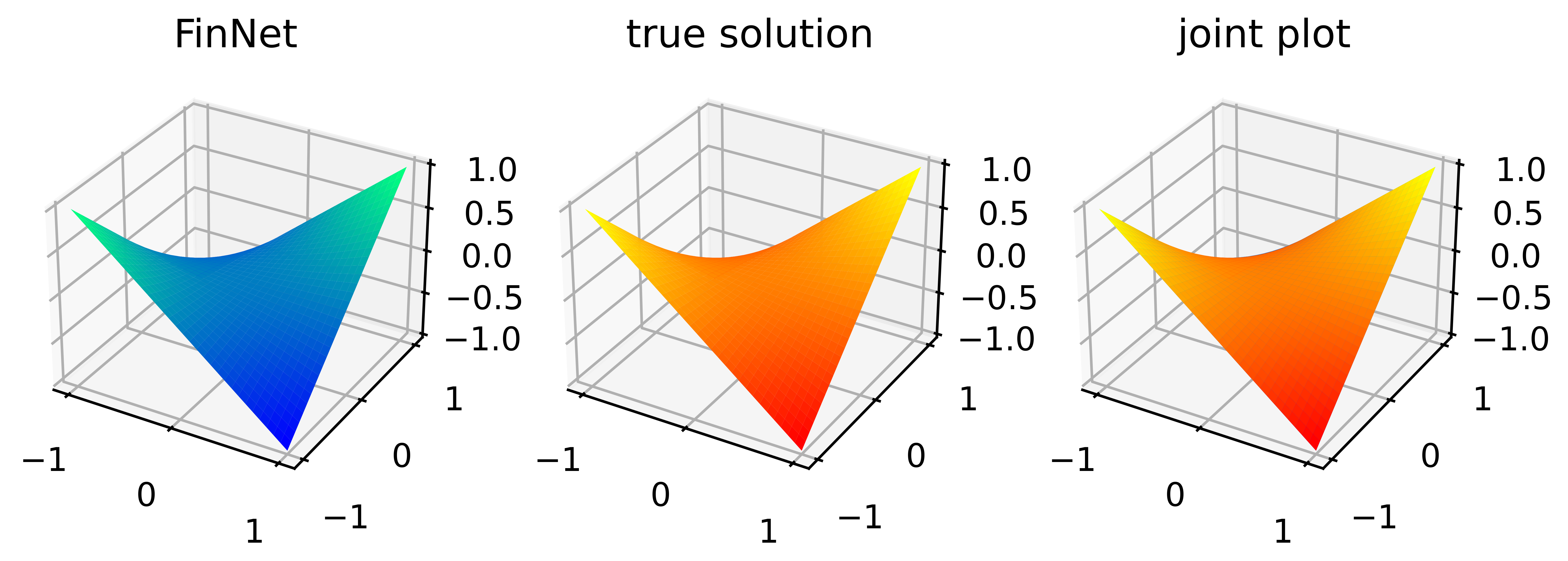}
    \caption{True solution versus the neural network's approximated solution for the Laplace equation (equation \ref{laplaceeq}).}
    \label{laplace}
\end{figure}

After $8,000$ epochs, the loss goes down to $0.088$, the MSE is between the true solution, and the predicted values is  $2.74\times 10^{-4}$. Note that the MSE between the true solution and the predicted values is much smaller than the loss of the neural network. This is reasonable since we are using finite difference to estimate the derivatives using a relatively coarse mesh grid with $N=32$. The plot of the true solution versus the neural network's approximated solution is as shown in figure \ref{laplace}.

\subsection{Example 5: Eikonal equation in two dimensions}

An Eikonal equation is a non-linear partial differential equation of first-order, which is commonly encountered in problems of wave propagation. Let $\Omega =  (-1,1)\times (-1,1)$, consider the equation 
\begin{equation}\label{eikonaleq}
\begin{cases}
|Du(x,y)|= 1 + \epsilon \Delta (x,y), & \text{for}\; (x,y)\in (-1,1)\times (-1,1),\\
u(x,y) = 1 - \sqrt{x^2+y^2} , &\text{for}\; (x,y)\in \partial \Omega.
\end{cases}    
\end{equation}

Here, we use $\epsilon= 0.0001$. The exact solution is 
$u^*(x,y) = 1 - \sqrt{x^2+y^2}.$

We used a neural network of four hidden layers with 64 neurons per layer and hyperbolic tangent activation functions to approximate the true solution. To learn the parameters, we use the Adam optimizer with learning rate $0.001$. The mesh size used is $N=32$. Then $G$ is formed by the grid $[-1,\frac{-29}{31},...,\frac{29}{31},1]\times [-1,\frac{-29}{31},...,\frac{29}{31},1]$ and $B$ consists of boundary grid points $(-1, -1), (-1, \frac{-29}{31}),..., (-1,\frac{29}{31}),(-1,1)$ and $(1, -1), (1, \frac{-29}{31}),..., (1,\frac{29}{31}),(1,1)$. 

Following the FinNet strategy, we train the network $v$ as follows: For each epoch, we first compute $\hat{u} \gets v(G)$, which also gives $\hat{u}_B \gets v(B)$. Then, we initialize $L \gets MSE(\hat{u}_B, g(B))$ to enforce the constraint $u(x,y) = 1 - \sqrt{x^2+y^2}$ on $\partial \Omega$ on the neural network. Next, we update the boundary values of $\hat{u}$ with the already known exact values based on $u(x,y) = 1 - \sqrt{x^2+y^2}$ for $(x,y)\in B$. Based on this newly updated $\hat{u}$, we estimate $|Du(x,y)|$ by finite difference. Then, we update the loss:
\begin{equation}
    L \gets L + \frac{1}{30^2}\sum_{i=1}^{30}\sum_{j=1}^{30}\Big(|D\hat{u}(x_i,y_j)|-1 - \epsilon \Delta(x_i,y_j)\Big)^2.
\end{equation}
where $x_1 = \frac{-29}{31}, ..., x_{30} =  \frac{29}{31}$ and $y_1 = \frac{-29}{31}, ..., y_{30} =  \frac{29}{31}$. After getting the loss, we update the weights of the neural network $v$.

\begin{figure}[htbp]
\centering
\includegraphics[scale=0.8]{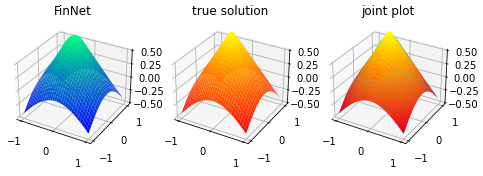}
    \caption{True solution versus the neural network's approximated solution for Eikonal equation (equation \ref{eikonaleq}).}
    \label{figeikonal}
\end{figure}

After $5,000$ epochs, the loss goes down to $0.01$, the MSE is between the true solution and the predicted value is  $7.40\times 10^{-5}$. The plot of the true solution versus the neural network's approximated solution is as shown in figure \ref{figeikonal}.

\section{Discussion and Conclusions}\label{sec:concl}
In this work, we analyzed potential issues when applying PINNs for differential equations and introduced a novel technique, namely FinNet, for solving differential equations by incorporating finite difference into deep learning. Even though the training phase is mesh-dependent, the prediction phase is mesh-free. We illustrated the effectiveness of our methods through experiments on solving various equations, which shows that the approximation provided by FinNet is very close to the true solution in terms of the MSE and may work even when PINNs do not.

For future work, various questions remain that are interesting to be addressed. Those can be questions on the hyperparameters for FinNet, such as how to choose the number of layers, activation function and mesh grid size. Furthermore, it would be interesting to compare FinNet with other approaches for nonlinear time-dependent PDEs or high-dimensional PDEs such as the high-dimensional Hamilton–Jacobi–Bellman equation, or the Burger's equation.

\bibliographystyle{unsrt}
\bibliography{bib.bib}

\end{document}